\documentclass[11pt,a4paper]{article}
\usepackage[hyperref]{acl2018}
\usepackage{times}
\usepackage{latexsym}
\usepackage{stfloats}
\usepackage{url}
\usepackage{float}
\usepackage{graphicx}
\usepackage{epstopdf}
\usepackage{amsfonts}
\usepackage{amssymb}
\usepackage{fixltx2e}
\usepackage{amsmath}
\usepackage{verbatim}
\usepackage{multirow}
\usepackage{amsmath}
\usepackage[noend]{algorithmic}
\usepackage[ruled]{algorithm2e}
\usepackage{subfigure}
\usepackage{pifont}% http://ctan.org/pkg/pifont
\usepackage{array}
\usepackage{fp}
\usepackage{makecell}
\usepackage{flushend}
\usepackage{cases}
\usepackage{color}
\usepackage{bbm}
\newcommand{\thickhline}{\noalign{\hrule height 1pt}}
\aclfinalcopy
\def\aclpaperid{102} 

\title{Learning Matching Models with Weak Supervision for Response Selection in Retrieval-based Chatbots}
\author{
	Yu Wu$^\dag$, Wei Wu$^\ddag$, Zhoujun Li$^\dag$\thanks{~~~Corresponding Author}~, Ming Zhou$^\diamondsuit$~~~\\
	$^\dag$State Key Lab of Software Development Environment, Beihang University, Beijing, China  \\
		$^\dag$ Authors are supported by AdeptMind Scholarship \\
	$^ \diamondsuit$Microsoft Research, Beijing, China\\
	$^\ddag$~~~~Microsoft Corporation, Beijing, China\\
	\{wuyu,lizj\}@buaa.edu.cn \{wuwei,mingzhou\}@microsoft.com 
}
\date{}

\begin{document}
	\maketitle
	\begin{abstract}
		We propose a method that can leverage unlabeled data to learn a matching model for response selection in retrieval-based chatbots. The  method employs a sequence-to-sequence architecture (Seq2Seq) model as a weak annotator to judge the matching degree of unlabeled pairs, and then performs learning with both the weak signals and the unlabeled data. Experimental results on two public data sets indicate that matching models get  significant improvements when they are learned with the proposed method. 
	\end{abstract}
	
	\section{Introduction}
	Recently, more and more attention from both academia and industry is paying to building non-task-oriented chatbots that can naturally converse with humans on any open domain topics. Existing approaches can be categorized into generation-based methods \cite{DBLP:conf/acl/ShangLL15,vinyals2015neural,serban2015building,sordoni2015neural,xing2017topic,serban2017hierarchical, xing2017hierarchical} which synthesize a response with natural language generation techniques, and retrieval-based methods \cite{hu2014convolutional,lowe2015ubuntu,DBLP:conf/sigir/YanSW16,zhou2016multi,wu2017sequential} which select a response from a pre-built index.  In this work, we study response selection for retrieval-based chatbots, not only because retrieval-based methods can return fluent and informative responses, but also because they have been successfully applied to many real products such as the social-bot XiaoIce from Microsoft \cite{shum2018eliza} and the E-commerce assistant AliMe Assist from Alibaba Group \cite{li2017alime}. 
	
	A key step to response selection is measuring the matching degree between a response candidate and an input which is either a single message \cite{hu2014convolutional} or a conversational context consisting of multiple utterances \cite{wu2017sequential}. While existing research focuses on how to define a matching model with neural networks, little attention has been paid to how to learn such a model when few labeled data are available. In practice, because human labeling is expensive and exhausting, one cannot have large scale labeled data for model training. Thus, a common practice is to transform the matching problem to a classification problem with human responses as positive examples and randomly sampled ones as negative examples. This strategy, however, oversimplifies the learning problem, as most of the randomly sampled responses are either far from the semantics of the messages or the contexts, or they are false negatives which pollute the training data as noise.  As a result, there often exists a significant gap between the performance of a model in training and the same model in practice \cite{wang2015syntax,wu2017sequential}.\footnote{The model performs well on randomly sampled data, but badly on human labeled data.}
	
	We propose a new method that can effectively leverage unlabeled data for learning matching models. To simulate the real scenario of a retrieval-based chatbot, we construct an unlabeled data set by retrieving response candidates from an index. Then, we employ a weak annotator to provide matching signals for the unlabeled input-response pairs, and leverage the signals to supervise the learning of matching models. The weak annotator is pre-trained from large scale human-human conversations without any annotations, and thus a Seq2Seq model becomes a natural choice. Our approach is compatible with any matching models, and falls in a teacher-student framework \cite{hinton2015distilling} where the Seq2Seq model transfers the knowledge from human-human conversations to the learning process of the matching models. Broadly speaking, both of \cite{hinton2015distilling} and our work let a neural network supervise the learning of another network.
	  An advantage of our method is that it turns the hard zero-one labels in the existing learning paradigm to soft (weak) matching scores. Hence, the model can learn a large margin between a true response with a true negative example, and the semantic distance between a true response and a false negative example is short. Furthermore, due to the simulation of real scenario, harder examples can been seen in the training phase that makes the model more robust in the testing.

	We conduct experiments on two public data sets, and experimental results on both data sets indicate that models learned with our method can significantly outperform their counterparts learned with the random sampling strategy. 
	
	Our contributions include: (1) proposal of a new method that can leverage unlabeled data to learn matching models for retrieval-based chatbots; and (2) empirical verification of the effectiveness of the method on public data sets.   
	
	\section{Approach}
	
	\subsection{The Existing Learning Approach}
	Given a data set $\mathcal{D} = \{x_i,(y_{i,1},\ldots, y_{i,n})\}_{i=1}^N$ with $x_i$ a message or a conversational context and $y_{i,j}$ a response candidate of $x_i$, we aim to learn a matching model $\mathcal{M}(\cdot, \cdot)$ from $\mathcal{D}$. Thus, for any new pair $(x,y)$, $\mathcal{M}(x, y)$ measures the matching degree between $x$ and $y$.    
	
	To obtain a matching model, one has to deal with two problems: (1) how to define $\mathcal{M}(\cdot, \cdot)$; and (2) how to perform learning. Existing work focuses on Problem (1) where state-of-the-art methods include dual LSTM \cite{lowe2015ubuntu}, Multi-View LSTM \cite{zhou2016multi}, CNN \cite{DBLP:conf/sigir/YanSW16}, and Sequential Matching Network \cite{wu2017sequential}, but adopts a simple strategy for Problem (2): $\forall x_i$, a human response is designated as $y_{i,1}$ with a label $1$, and some randomly sampled responses are treated as $(y_{i,2},\ldots,y_{i,n})$ with labels $0$. $\mathcal{M}(\cdot,\cdot)$ is then learned by maximizing the following objective:
	\begin{equation}\label{oriobj}
	\small
	\resizebox{1\hsize}{!}{$
		\sum_{i=1}^{N} \sum_{j=1}^n \left[r_{i,j} \log(\mathcal{M}(x_i,y_{i,j})) + (1-r_{i,j})\log(1-\mathcal{M}(x_i,y_{i,j}))\right],$}
	\end{equation} 
	where $r_{i,j}\in \{0,1\}$ is a label. While matching accuracy can be improved by carefully designing $\mathcal{M}(\cdot,\cdot)$ \cite{wu2017sequential}, the bottleneck becomes the learning approach which suffers obvious problems: most of the randomly sampled $y_{i,j}$ are semantically far from $x_i$ which may cause an undesired decision boundary at the end of optimization; some $y_{i,j}$ are false negatives. As hard zero-one labels are adopted in Equation (\ref{oriobj}), these false negatives may mislead the learning algorithm.  The problems remind us that besides good architectures of matching models, we also need a good approach to learn such models from data.

	% Prior work mainly focuses on how to design the network architecture of $\mathcal{M}$, from simple models like  and MultiView LSTM \cite{zhou2016multi} to complex models like Deep Learning to Respond \cite{DBLP:conf/sigir/YanSW16} and Sequential Matching Network \cite{wu2016sequential}. In this paper, we target at exploiting the unlabeled data by leveraging a weak annotator $G$ to give a signal $s_i$ for each negative sampled instance $(x_i,y_j)$. Then the matching model $\mathcal{M}$ is trained with both $s_{i,j}$ and $(x_i,y_j), \forall i,j$, so that the knowledge of $G$ can transfer to the matching model.
	
	\subsection{A New Learning Method}
	As human labeling is infeasible when training complicated neural networks, we propose a new method that can leverage unlabeled data to learn a matching model. Specifically, instead of random sampling, we construct $\mathcal{D}$ by retrieving $(y_{i,2},\ldots,y_{i,n})$ from an index ($y_{i,1}$ is the human response of $x_i$). By this means, 
	some $y_{i,j}$ are true positives, and some are negatives but semantically close to $x_i$. After that, we employ a weak annotator $G(\cdot,\cdot)$ to indicate the matching degree of every $(x_i, y_{i,j})$ in $\mathcal{D}$ as weak supervision signals. Let $s_{ij} = G(x_i,y_{i,j})$, then the learning approach can be formulated as:
	\begin{equation} 	\small 
	\underset{\mathcal{M}(\cdot,\cdot)}{\arg\min} \sum_{i=1}^N \sum_{j=1}^n \max(0, \mathcal{M}(x_i,y_{i,j}) - \mathcal{M}(x_i,y_{i,1}) + s'_{i,j}),
	\label{loss}
	\end{equation}
	where $s'_{ij}$ is a normalized weak signal defined as $max(0,\frac{s_{i,j}}{s_{i,1}}-1)$. The normalization here eliminates bias from different $x_{i}$. 
	
  Objective (\ref{loss}) encourages a large margin between the matching of an input and its human response and the matching of the input and a negative response judged by $G(\cdot,\cdot)$ (as will be seen later, $\frac{s_{i,j}}{s_{i,1}}>1$). The learning approach simulates how we build a matching model in a retrieval-based chatbot: given $\{x_i\}$, some response candidates are first retrieved from an index. Then human annotators are hired to judge the matching degree of each pair. Finally, both the data and the human labels are fed to an optimization program for model training. Here, we replace the expensive human labels with cheap judgment from $G(\cdot,\cdot)$.
	
	We define $G(\cdot,\cdot)$ as a sequence-to-sequence architecture \cite{vinyals2015neural} with an attention mechanism \cite{bahdanau2014neural}, and pre-train it with large amounts of human-human conversation data. The Seq2Seq model can capture the semantic correspondence between an input and a response, and then transfer the knowledge to the learning of a matching model in the optimization of (\ref{loss}). $s_{ij}$ is then defined as the likelihood of generating $y_{i,j}$ from $x_i$: 
	\begin{equation} 	\small
	s_{ij} = \sum_{k} \log[p(w_{y_{i,j}, k}, | x_i, w_{y_{i,j}, l < k})],
	\end{equation}
	where $w_{y_{i,j}, k}$ is the $k$-th word of $y_{i,j}$ and $w_{y_{i,j}, l < k}$ is the word sequence before $w_{y_{i,j}, k}$.
	
%	Equation (\ref{loss}) turns the hard zero-one labels in Equation (\ref{oriobj}) to soft matching degrees, and thus some false negatives in the existing learning approach can be rectified ($\frac{s_{i,j}}{s_{i,1}}<1$), and different levels of true negatives refer to different margins. Our method encourages the model to be more confident about the true response when it has a lower $s_{i,1}$ than a negative example. 
	Since negative examples are retrieved by a search engine, the oversimplification problem of the negative sampling approach can be partially mitigated. We leverage a weak annotator to assign a score for each example to distinguish false negative examples and true negative examples.  Equation (\ref{loss}) turns the hard zero-one labels in Equation (\ref{oriobj}) to soft matching degrees, and thus our method encourages the model to be more confident to classify a response with a high $s_{i,j}$ score as a negative one. In this way, we can avoid false negative examples and true negative examples are treated equally during training, and update the model toward a correct direction.  
	
	It is noteworthy that although our approach also involves an interaction between a generator and a discriminator, it is different from the GANs \cite{goodfellow2014generative} in principle. GANs try to learn a better generator via an adversarial process, while our approach aims to improve the discriminator with supervision from the generator, which also differentiates it from the recent work on transferring knowledge from a discriminator to a generative visual dialog model \cite{lu2017best}. Our approach is also different from those semi-supervised approaches in the teacher-student framework \cite{dehghani2017fidelity,dehghani2017avoiding}, as there are no labeled data  in learning. 

	\section{Experiment}
	We conduct experiments on two public data sets: STC data set \cite{wang2013dataset} for single-turn response selection and  Douban Conversation Corpus \cite{wu2017sequential} for multi-turn response selection. Note that we do not test the proposed approach on Ubuntu Corpus \cite{lowe2015ubuntu}, because both training and test data in the corpus are constructed by random sampling. %The test sets are labeled by human, but the training data sets do not contain annotated negative instances. 

	\subsection{Implementation Details}
	We implement our approach with TensorFlow. In both experiments, the same Seq2Seq model is exploited which is trained with $3.3$ million input-response pairs extracted from the training set of the Douban data. Each input is a concatenation of consecutive utterances in a context, and the response is the next turn ($\{u_{<i}\},u_i$).  We set the vocabulary size as $30,000$, the hidden vector size as $1024$, and the embedding size as $620$.  Optimization is conducted with stochastic gradient descent \cite{bottou2010large}, and is terminated when perplexity on a validation set ($170$k pairs) does not decrease in $3$ consecutive epochs. In optimization of Objective (\ref{loss}), we initialize $\mathcal{M}(\cdot,\cdot)$ with a model trained under Objective (\ref{oriobj}) with the (random) negative sampling strategy, and fix word embeddings throughout training. This can stabilize the learning process. The learning rate is fixed as $0.1$.

	%which contains $0.5$ million conversation sessions. We extract adjacent utterances in the sessions as parallel texts to train the Seq2Seq model. We have 3.34 million pairs for training and 168,784 pairs for validation. 
	\subsection{Single-turn Response Selection}
	
	\textbf{Experiment settings}: in the STC (stands for Short Text Conversation) data set, the task is to select a proper response for a post in Weibo\footnote{\url{http://weibo.sina.com}}. The training set contains $4.8$ million post-response (true response) pairs. The test set consists of $422$ posts with each one associated with around $30$ responses labeled by human annotators in ``good'' and ``bad''. In total, there are $12,402$ labeled pairs in the test data. Following \cite{wang2013dataset, wang2015syntax}, we combine the score from a matching model with TF-IDF based cosine similarity using RankSVM whose parameters are chosen by $5$-fold cross validation. Precision at position 1 (P@1) is employed as an evaluation metric. In addition to the models compared on the data in the existing literatures, we also implement dual LSTM \cite{lowe2015ubuntu} as a baseline. As case studies, we learn a dual LSTM and an CNN \cite{hu2014convolutional} with the proposed approach, and denote them  as LSTM+WS (Weak Supervision) and CNN+WS, respectively. When constructing  $\mathcal{D}$, we build an index with the training data using Lucene\footnote{\url{https://lucenenet.apache.org/}} and retrieve $9$ candidates (i.e., $\{y_{i,2},\ldots,y_{i,n}\}$) for each post with the inline algorithm of the index. We form a validation set by randomly sampling $10$ thousand posts associated with the responses from $\mathcal{D}$ (human response is positive and others are treated as negative).

	\begin{table}[h]
		\small
	
		\centering
		\begin{tabular}{l|c}
			\thickhline
			& P@1  \\ \hline
			TFIDF  \cite{wang2013dataset} & 0.574\\ 
			+Translation  \cite{wang2013dataset} & 0.587\\ 
			+WordEmbedding & 0.579\\ 
			+DeepMatch$_{topic}$ \cite{lu2013deep} & 0.587 \\
			+DeepMatch$_{tree}$ \cite{wang2015syntax} & 0.608\\  \hline			
			+LSTM \cite{lowe2015ubuntu} & 0.592 \\ 
			+LSTM+WS & 0.616 \\ \hline			
			+CNN \cite{hu2014convolutional} & 0.585\\ 
			+CNN+WS & 0.604\\ 
			\thickhline
		\end{tabular}	\caption{Results on STC \label{exp:single}}
	\end{table}
	
	\textbf{Results}: Table \ref{exp:single} reports the results. We can see that CNN and LSTM consistently get improved when learned with the proposed approach, and the improvements over the models learned with random sampling are statistically significant (t-test with $p$-value $< 0.01$). LSTM+WS even surpasses the best performing model, DeepMatch$_{tree}$, reported on this data. These results indicate the usefulness of the proposed approach in practice. One can expect improvements to models like DeepMatch$_{tree}$ with the new learning method. We leave the verification as future work.

	\subsection{Multi-turn Response Selection}
	
	\textbf{Experiment settings}:  Douban Conversation Corpus contains $0.5$ million context-response (true response) pairs for training and 1000 contexts for test. In the test set, every context has 10 response candidates, and each of the response has a label ``good" or ``bad" judged by human annotators. Mean average precision (MAP) \cite{baeza1999modern}, mean reciprocal rank (MRR) \cite{voorhees1999trec}, and precision at position 1 (P@1)  are employed as evaluation metrics. We copy the numbers reported in \cite{wu2017sequential} for the baseline models, and  
	learn LSTM, Multi-View, and SMN with the proposed approach.  We build an index with the training data, and retrieve $9$ candidates with the method in \cite{wu2017sequential} for each context when constructing $\mathcal{D}$. $10$ thousand pairs are sampled from $\mathcal{D}$ as a validation set. 
	
	\textbf{Results}: Table \ref{exp:multi} reports the results. Consistent with the results on the STC data, every model (+WS one) gets improved with the new learning approach, and the improvements are statistically significant (t-test with $p$-value $< 0.01$).

	\begin{table}[h]
		\small
		
		\centering
		\begin{tabular}{l|c|c|c}
			\thickhline
			&MAP&MRR& P@1  \\ \hline
			TFIDF  & 0.331 &0.359 &0.180\\ 
			RNN  & 0.390 &0.422 &0.208\\ 
			CNN & 0.417 &0.440 &0.226\\ 
			
			BiLSTM &0.479&0.514&0.313\\ 	
			DL2R \cite{DBLP:conf/sigir/YanSW16} &0.488&0.527&0.330 \\ 	 \hline	
			LSTM \cite{lowe2015ubuntu} & 0.485 & 0.527 &0.320 \\
			LSTM+WS & 0.519 & 0.559 &0.359 \\ \hline	
			Multi-View \cite{zhou2016multi} &0.505&0.543&0.342 \\ 
			Multi-View+WS &0.534&0.575&0.378 \\ 	\hline
			
			SMN \cite{wu2017sequential} &0.526&0.571&0.393\\ 
			SMN+WS &0.565&0.609&0.421\\ 
			\thickhline
		\end{tabular}	\caption{Results on Douban Conversation Corpus \label{exp:multi}}
	\end{table}

	\subsection{Discussion}
	\textbf{Ablation studies}: we first replace the weak supervision $s_{i,j}'$ in Equation (\ref{loss}) with a constant $\epsilon$ selected from $\{0.1,0.2, \ldots, 0.9\}$ on validation, and denote the models as model+const. Then, we keep everything the same as our approach but replace $\mathcal{D}$ with a set constructed by random sampling, denoted as model+WSrand. Table \ref{exp:abl} reports the results. We can conclude that both the weak supervision and the strategy of training data construction are important to the success of the proposed learning approach. Training data construction plays a more crucial role, because it involves more true positives and negatives with different semantic distances to the positives into learning.

	\begin{table}[t]
		\small
		
		\centering
		\begin{tabular}{l|c|c|c|c}
			\thickhline
			&	\multicolumn{1}{c|}{STC} & \multicolumn{3}{c}{Douban}\\ \hline
			& P@1 &MAP&MRR& P@1  \\ \hline
			CNN+WSrand &0.590&-&-&-  \\
			CNN+const  &0.598&-&-&- \\
			CNN+WS &0.604&-&-&- \\ \hline
			LSTM+WSrand &0.598& 0.501& 0.532&0.323  \\
			LSTM+const &0.607 & 0.510 & 0.545 &0.331 \\
			LSTM+WS&0.616 & 0.519 & 0.559 &0.359 \\ \hline
			Multi-View+WSrand &-&0.515&0.549&0.357 \\	
			Multi-View+const &- &0.528&0.564&0.370 \\ 
			Multi-View+WS&- &0.534&0.575&0.378 \\ 	\hline
			SMN+WSrand &- & 0.536 & 0.574 &0.377 \\
			SMN+const &- &0.558&0.603&0.417\\ 
			SMN+WS &-&0.565&0.609&0.421\\ 
			\thickhline
		\end{tabular}	\caption{Ablation results. \label{exp:abl}}
	\end{table}

	\textbf{Does updating the Seq2Seq model help?} It is well known that Seq2Seq models suffer from the ``safe response'' \cite{li2015diversity} problem, which may bias the weak supervision signals to high-frequency responses. Therefore, we attempt to iteratively optimize the Seq2Seq model and the matching model and check if the matching model can be further improved. Specifically, we update the Seq2Seq model every $20$ mini-batches with the policy-based reinforcement learning approach proposed in \cite{li2016deep}. The reward is defined as the matching score of a context and a response given by the matching model. Unfortunately, we do not observe significant improvement on the matching model. The result is attributed to two factors: (1) it is difficult to significantly improve the Seq2Seq model with a policy gradient based method; and (2) eliminating ``safe response" for Seq2Seq model cannot help a matching model to learn a better decision boundary.

	\textbf{How the number of response candidates affects learning}: we vary the number of $\{y_{i,j}\}_{j=1}^n$ in $\mathcal{D}$ in $\{2,5,10,20\}$ and study how the hyper-parameter influences learning. We study with LSTM on the STC data and SMN on the Douban data.  Table \ref{exp:instance_number} reports the results. We can see that as the number of candidates increases, the performance of the the learned models becomes better.  Even with $2$ candidates (one from human and the other from retrieval), our approach can still improve the peformance of matching models. 
	
	\begin{table}[h]
		\small
		
		\centering
		\begin{tabular}{l|c|c|c|c}
			\thickhline
			&LSTM$_2$ &LSTM$_5$&LSTM$_{10}$& LSTM$_{20}$  \\ \hline
			P@1&0.603&0.608&0.615&0.616 \\ \hline
			&SMN$_2$ &SMN$_5$&SMN$_{10}$& SMN$_{20}$  \\ \hline
			MAP&0.542&0.556&0.565&0.567\\\hline
			MRR&0.588&0.594&0.609&0.609\\\hline
			P@1&0.408&0.412&0.421&0.423\\\hline
			\thickhline
		\end{tabular}	\caption{The effect of instance number \label{exp:instance_number}}
	\end{table}
	%	, which is depicted in Figure \ref{fig:performance}. We select LSTM on the STC dataset and SMN on Douban Conversation dataset to illustrate the tendency. We can see that the performance does not increase when the number reaches 10.
	%		\begin{figure}[h]		
	%		\begin{center} 
	%			\includegraphics[width=8cm,height=3.5cm]{acl_negins.pdf}
	%		\end{center}
	%		
	%		\caption{Performance in terms of different numbers on retrieved instances.}\label{fig:performance}
	%	\end{figure}
	
	\section{Conclusion and Future Work}
	Previous studies focus on architecture design for retrieval-based chatbots, but neglect the problems brought by random negative sampling in the learning process. 
	In this paper, we propose leveraging a Seq2Seq model as a weak annotator on unlabeled data to learn a matching model for response selection. By this means, we can mine hard instances for matching model and give them scores with a weak annotator. Experimental results on public data sets verify the effectiveness of the new learning approach. In the future, we will investigate how to remove bias from the weak supervisors, and further improve the matching model performance with a semi-supervised approach. 
	
	\section*{Acknowledgment}
Yu Wu is supported by Microsoft Fellowship Scholarship
and AdeptMind Scholarship.	This work is supported by the National Natural Science Foundation of China (Grand Nos. 61672081,U1636211,61370126), Beijing Advanced Innovation Center for Imaging Technology (No.BAICIT-2016001). 
	\bibliography{acl2018}

\begin{thebibliography}{}
\expandafter\ifx\csname natexlab\endcsname\relax\def\natexlab#1{#1}\fi

\bibitem[{Baeza-Yates et~al.(1999)Baeza-Yates, Ribeiro-Neto
  et~al.}]{baeza1999modern}
Ricardo Baeza-Yates, Berthier Ribeiro-Neto, et~al. 1999.
\newblock {\em Modern information retrieval\/}, volume 463.
\newblock ACM press New York.

\bibitem[{Bahdanau et~al.(2015)Bahdanau, Cho, and Bengio}]{bahdanau2014neural}
Dzmitry Bahdanau, Kyunghyun Cho, and Yoshua Bengio. 2015.
\newblock Neural machine translation by jointly learning to align and
  translate.
\newblock {\em ICLR\/} .

\bibitem[{Bottou(2010)}]{bottou2010large}
L{\'e}on Bottou. 2010.
\newblock Large-scale machine learning with stochastic gradient descent.
\newblock In {\em Proceedings of COMPSTAT'2010\/}, Springer, pages 177--186.

\bibitem[{Dehghani et~al.(2017{\natexlab{a}})Dehghani, Mehrjou, Gouws, Kamps,
  and Sch{\"o}lkopf}]{dehghani2017fidelity}
Mostafa Dehghani, Arash Mehrjou, Stephan Gouws, Jaap Kamps, and Bernhard
  Sch{\"o}lkopf. 2017{\natexlab{a}}.
\newblock Fidelity-weighted learning.
\newblock {\em arXiv preprint arXiv:1711.02799\/} .

\bibitem[{Dehghani et~al.(2017{\natexlab{b}})Dehghani, Severyn, Rothe, and
  Kamps}]{dehghani2017avoiding}
Mostafa Dehghani, Aliaksei Severyn, Sascha Rothe, and Jaap Kamps.
  2017{\natexlab{b}}.
\newblock Avoiding your teacher's mistakes: Training neural networks with
  controlled weak supervision.
\newblock {\em arXiv preprint arXiv:1711.00313\/} .

\bibitem[{Goodfellow et~al.(2014)Goodfellow, Pouget-Abadie, Mirza, Xu,
  Warde-Farley, Ozair, Courville, and Bengio}]{goodfellow2014generative}
Ian Goodfellow, Jean Pouget-Abadie, Mehdi Mirza, Bing Xu, David Warde-Farley,
  Sherjil Ozair, Aaron Courville, and Yoshua Bengio. 2014.
\newblock Generative adversarial nets.
\newblock In {\em Advances in neural information processing systems\/}. pages
  2672--2680.

\bibitem[{Hinton et~al.(2015)Hinton, Vinyals, and Dean}]{hinton2015distilling}
Geoffrey Hinton, Oriol Vinyals, and Jeff Dean. 2015.
\newblock Distilling the knowledge in a neural network.
\newblock {\em arXiv preprint arXiv:1503.02531\/} .

\bibitem[{Hu et~al.(2014)Hu, Lu, Li, and Chen}]{hu2014convolutional}
Baotian Hu, Zhengdong Lu, Hang Li, and Qingcai Chen. 2014.
\newblock Convolutional neural network architectures for matching natural
  language sentences.
\newblock In {\em Advances in Neural Information Processing Systems\/}. pages
  2042--2050.

\bibitem[{Li et~al.(2017)Li, Qiu, Chen, Wang, Gao, Huang, Ren, Zhao, Zhao, Wang
  et~al.}]{li2017alime}
Feng-Lin Li, Minghui Qiu, Haiqing Chen, Xiongwei Wang, Xing Gao, Jun Huang,
  Juwei Ren, Zhongzhou Zhao, Weipeng Zhao, Lei Wang, et~al. 2017.
\newblock Alime assist: An intelligent assistant for creating an innovative
  e-commerce experience.
\newblock In {\em Proceedings of the 2017 ACM on Conference on Information and
  Knowledge Management\/}. ACM, pages 2495--2498.

\bibitem[{Li et~al.(2016{\natexlab{a}})Li, Galley, Brockett, Gao, and
  Dolan}]{li2015diversity}
Jiwei Li, Michel Galley, Chris Brockett, Jianfeng Gao, and Bill Dolan.
  2016{\natexlab{a}}.
\newblock A diversity-promoting objective function for neural conversation
  models.
\newblock In {\em {NAACL} {HLT} 2016, The 2016 Conference of the North American
  Chapter of the Association for Computational Linguistics: Human Language
  Technologies, San Diego California, USA, June 12-17, 2016\/}. pages 110--119.

\bibitem[{Li et~al.(2016{\natexlab{b}})Li, Monroe, Ritter, Jurafsky, Galley,
  and Gao}]{li2016deep}
Jiwei Li, Will Monroe, Alan Ritter, Dan Jurafsky, Michel Galley, and Jianfeng
  Gao. 2016{\natexlab{b}}.
\newblock Deep reinforcement learning for dialogue generation.
\newblock In {\em Proceedings of the 2016 Conference on Empirical Methods in
  Natural Language Processing, {EMNLP} 2016, Austin, Texas, USA, November 1-4,
  2016\/}. pages 1192--1202.

\bibitem[{Lowe et~al.(2015)Lowe, Pow, Serban, and Pineau}]{lowe2015ubuntu}
Ryan Lowe, Nissan Pow, Iulian Serban, and Joelle Pineau. 2015.
\newblock The ubuntu dialogue corpus: A large dataset for research in
  unstructured multi-turn dialogue systems.
\newblock {\em SIGDIAL\/} .

\bibitem[{Lu et~al.(2017)Lu, Kannan, Yang, Parikh, and Batra}]{lu2017best}
Jiasen Lu, Anitha Kannan, Jianwei Yang, Devi Parikh, and Dhruv Batra. 2017.
\newblock Best of both worlds: Transferring knowledge from discriminative
  learning to a generative visual dialog model.
\newblock In {\em Advances in Neural Information Processing Systems\/}. pages
  313--323.

\bibitem[{Lu and Li(2013)}]{lu2013deep}
Zhengdong Lu and Hang Li. 2013.
\newblock A deep architecture for matching short texts.
\newblock In {\em Advances in Neural Information Processing Systems\/}. pages
  1367--1375.

\bibitem[{Serban et~al.(2016)Serban, Sordoni, Bengio, Courville, and
  Pineau}]{serban2015building}
Iulian~Vlad Serban, Alessandro Sordoni, Yoshua Bengio, Aaron~C. Courville, and
  Joelle Pineau. 2016.
\newblock End-to-end dialogue systems using generative hierarchical neural
  network models.
\newblock In {\em Proceedings of the Thirtieth {AAAI} Conference on Artificial
  Intelligence, February 12-17, 2016, Phoenix, Arizona, {USA.}\/}. pages
  3776--3784.

\bibitem[{Serban et~al.(2017)Serban, Sordoni, Lowe, Charlin, Pineau, Courville,
  and Bengio}]{serban2017hierarchical}
Iulian~Vlad Serban, Alessandro Sordoni, Ryan Lowe, Laurent Charlin, Joelle
  Pineau, Aaron~C Courville, and Yoshua Bengio. 2017.
\newblock A hierarchical latent variable encoder-decoder model for generating
  dialogues.
\newblock In {\em AAAI\/}. pages 3295--3301.

\bibitem[{Shang et~al.(2015)Shang, Lu, and Li}]{DBLP:conf/acl/ShangLL15}
Lifeng Shang, Zhengdong Lu, and Hang Li. 2015.
\newblock Neural responding machine for short-text conversation.
\newblock In {\em {ACL} 2015, July 26-31, 2015, Beijing, China, Volume 1: Long
  Papers\/}. pages 1577--1586.

\bibitem[{Shum et~al.(2018)Shum, He, and Li}]{shum2018eliza}
Heung-Yeung Shum, Xiaodong He, and Di~Li. 2018.
\newblock From eliza to xiaoice: Challenges and opportunities with social
  chatbots.
\newblock {\em arXiv preprint arXiv:1801.01957\/} .

\bibitem[{Sordoni et~al.(2015)Sordoni, Galley, Auli, Brockett, Ji, Mitchell,
  Nie, Gao, and Dolan}]{sordoni2015neural}
Alessandro Sordoni, Michel Galley, Michael Auli, Chris Brockett, Yangfeng Ji,
  Margaret Mitchell, Jian{-}Yun Nie, Jianfeng Gao, and Bill Dolan. 2015.
\newblock A neural network approach to context-sensitive generation of
  conversational responses.
\newblock In {\em {NAACL} {HLT} 2015, The 2015 Conference of the North American
  Chapter of the Association for Computational Linguistics: Human Language
  Technologies, Denver, Colorado, USA, May 31 - June 5, 2015\/}. pages
  196--205.

\bibitem[{Vinyals and Le(2015)}]{vinyals2015neural}
Oriol Vinyals and Quoc Le. 2015.
\newblock A neural conversational model.
\newblock {\em arXiv preprint arXiv:1506.05869\/} .

\bibitem[{Voorhees(1999)}]{voorhees1999trec}
Ellen~M. Voorhees. 1999.
\newblock The {TREC-8} question answering track report.
\newblock In {\em Proceedings of The Eighth Text REtrieval Conference, {TREC}
  1999, Gaithersburg, Maryland, USA, November 17-19, 1999\/}.

\bibitem[{Wang et~al.(2013)Wang, Lu, Li, and Chen}]{wang2013dataset}
Hao Wang, Zhengdong Lu, Hang Li, and Enhong Chen. 2013.
\newblock A dataset for research on short-text conversations.
\newblock In {\em Proceedings of the 2013 Conference on Empirical Methods in
  Natural Language Processing, {EMNLP} 2013, 18-21 October 2013, Grand Hyatt
  Seattle, Seattle, Washington, USA, {A} meeting of SIGDAT, a Special Interest
  Group of the {ACL}\/}. pages 935--945.

\bibitem[{Wang et~al.(2015)Wang, Lu, Li, and Liu}]{wang2015syntax}
Mingxuan Wang, Zhengdong Lu, Hang Li, and Qun Liu. 2015.
\newblock Syntax-based deep matching of short texts.
\newblock In {\em Twenty-Fourth International Joint Conference on Artificial
  Intelligence\/}.

\bibitem[{Wu et~al.(2017)Wu, Wu, Xing, Zhou, and Li}]{wu2017sequential}
Yu~Wu, Wei Wu, Chen Xing, Ming Zhou, and Zhoujun Li. 2017.
\newblock Sequential matching network: A new architecture for multi-turn
  response selection in retrieval-based chatbots.
\newblock In {\em Proceedings of the 55th Annual Meeting of the Association for
  Computational Linguistics (Volume 1: Long Papers)\/}. volume~1, pages
  496--505.

\bibitem[{Xing et~al.(2017)Xing, Wu, Wu, Liu, Huang, Zhou, and
  Ma}]{xing2017topic}
Chen Xing, Wei Wu, Yu~Wu, Jie Liu, Yalou Huang, Ming Zhou, and Wei-Ying Ma.
  2017.
\newblock Topic aware neural response generation.
\newblock In {\em AAAI 2017\/}. pages 3351--3357.

\bibitem[{Xing et~al.(2018)Xing, Wu, Wu, Zhou, Huang, and
  Ma}]{xing2017hierarchical}
Chen Xing, Wei Wu, Yu~Wu, Ming Zhou, Yalou Huang, and Wei-Ying Ma. 2018.
\newblock Hierarchical recurrent attention network for response generation.
\newblock {\em AAAI-18\/} .

\bibitem[{Yan et~al.(2016)Yan, Song, and Wu}]{DBLP:conf/sigir/YanSW16}
Rui Yan, Yiping Song, and Hua Wu. 2016.
\newblock Learning to respond with deep neural networks for retrieval-based
  human-computer conversation system.
\newblock In {\em Proceedings of the 39th International {ACM} {SIGIR}
  conference on Research and Development in Information Retrieval, {SIGIR}
  2016, Pisa, Italy, July 17-21, 2016\/}. pages 55--64.

\bibitem[{Zhou et~al.(2016)Zhou, Dong, Wu, Zhao, Yu, Tian, Liu, and
  Yan}]{zhou2016multi}
Xiangyang Zhou, Daxiang Dong, Hua Wu, Shiqi Zhao, Dianhai Yu, Hao Tian, Xuan
  Liu, and Rui Yan. 2016.
\newblock Multi-view response selection for human-computer conversation.
\newblock In {\em Proceedings of the 2016 Conference on Empirical Methods in
  Natural Language Processing, {EMNLP} 2016, Austin, Texas, USA, November 1-4,
  2016\/}. pages 372--381.

\end{thebibliography}
	\bibliographystyle{acl_natbib}
\end{document}